\documentclass[11pt]{article}
\usepackage{coling2020}
\usepackage{times}
\usepackage{url}
\usepackage{latexsym}
\usepackage{enumitem}
\usepackage{hyperref}
\usepackage{makecell}
\usepackage{longtable}
\usepackage{multirow}
\usepackage{etoolbox}
\usepackage{graphicx}

\usepackage[export]{adjustbox}
\usepackage[title]{appendix}
\usepackage{etoolbox}
\patchcmd{\appendices}{\quad}{. }{}{}

\title{
Financial Document Causality Detection Shared Task (FinCausal 2020)}

 \author{
\begin{tabular}[t]{ccc}
Dominique Mariko$^{1}$ & Hanna Abi-Akl $^{1}$ & Estelle Labidurie$^{1}$ \\ Stéphane Durfort$^{1}$ & Hugues de Mazancourt$^{1}$  & Mahmoud El-Haj$^{2}$\\
\\
\end{tabular}
\\
$^{1}$YseopLab, FR,
$^{2}$Lancaster University, UK, \\
$^{1}$\tt{lab@yseop.com}, 
$^{2}$\tt{m.el-haj@lancaster.ac.uk} \\
}

\date{12/12/2020}

\begin{document}

\maketitle

\begin{abstract}
We present the FinCausal 2020 Shared Task on Causality Detection in Financial Documents and the associated FinCausal dataset, and discuss the participating systems and results. 
Two sub-tasks are proposed: a binary classification task (Task 1) and a relation extraction task (Task 2). A total of 16 teams submitted runs across the two Tasks and 13 of them contributed with a system description paper.
This workshop is associated to the Joint Workshop on Financial Narrative Processing and MultiLing Financial Summarisation (FNP-FNS 2020\footnote{\url{http://wp.lancs.ac.uk/cfie/fnp2020/}}), held at The 28th International Conference on Computational Linguistics (COLING'2020\footnote{\url{https://coling2020.org}}), Barcelona, Spain on September 12, 2020.
\end{abstract}

\section{Introduction}
\label{overview}

In an effort to automatically interpret the semantics of written languages, the analysis and understanding of causal relationships between facts stand as a key element. A major difficulty regarding automation is that causality can be expressed using many different syntactic patterns as well as contrasted semantic representations. This difficulty is reinforced by the existence of both explicit and implicit cause-effect links. Early works in this field, such as \cite{Kho:20}, aim at detecting causal relations using linguistic patterns. However, these applications are often restricted to a specific domain, limited to explicit causal relationships only (causal links, causative verbs, resultative constructions, conditionals and causative adverbs and adjectives), and do not take into account the ambiguities of the connectors. 
The semi-automatic method developed by \cite{Gir:21} goes a step further by creating lexico-syntactic patterns based on WordNet semantic relations between nouns, then using semantics constraints to test ambiguous causal relations.
Despite better results, the exclusive use of linguistic patterns prevents a fully efficient coverage of all cause-effect links. Consequently, various machine learning techniques were tested for this task. \cite{Cha:22} developed Naive Bayes causality extraction models based on lexical pair probability and cue phrase probability. 
By focusing on the dynamics of causal relationships, the system PREPOST developed by \cite{Sil:23} stands as a viable system to detect causal relationships between one event and a consequent state of this event, training a classifier to identify events' preconditions and/or postconditions.
In parallel, hybrid methods were also developed such as the expanded semantic parsing of \cite{Dun:18}. This system combines an SCL approach, pattern-based methods and a neural network architecture, offering more flexibility than exclusive pattern based approaches.
Overall, the management of linguistic ambiguities as well as the existence of implicit connections appear to be the main brakes in the identification and extraction of causal relationships. 

In this paper, we present the FinCausal Corpus and the associated featured Tasks, as a contribution to the research effort addressing implicit and multiple causalities detection automation in financial documents. All the  datasets  created  for  this  shared  task are publicly available to support further research on Causality modelling\footnote{\url{https://competitions.codalab.org/competitions/25340}}, and the detailed annotation scheme is provided in the Appendix A.\blfootnote{This work is licensed under a Creative Commons Attribution 4.0 International License. License details: \url{http://creativecommons.org/licenses/by/4.0/}.}
Next, Section 2 describes the FinCausal Corpus and Section 3 presents the Tasks. Section 4 provides the baseline proposed to participants and details their results, with  a  high-level  description  of  the  approaches they  adopted. Finally, Section 5 concludes this report and discusses some future directions.

\section{FinCausal Corpus}

The data are extracted from a corpus of 2019 financial news provided by \href{http://www.qwamci.com/}{Qwam}, collected on 14.000 economics and finance websites.
The original raw corpus is an ensemble of HTML pages corresponding to daily information retrieval from financial news feed. These news mostly inform on the 2019 financial landscape, but can also contain information related to politics, micro economics or other topic considered relevant for finance information. 
Data are released under the CC0 License\footnote{\url{https://creativecommons.org/publicdomain/zero/1.0/deed.en}}.\\

All collected HTML files were initially split into sentences according to their punctuation, then were grouped into text sections of 1 to 3 sentences after the data annotation process has been completed.
Below are the principle global metrics gathered during the annotation process. The metrics are defined with respect to the annotation scheme. 

For consistency in our references, we refer to a \textbf{file} as the original document to process, a \textbf{text section} as a multi-sentenced text string (1 to 3 sentences that may or may not contain causality) and a \textbf{chunk} as a substring (consisting either of a part of a sentence, a whole sentence or multiple sentences) within a text section. We also retain statistics related to the main tags used in our annotation scheme during the preparation of the datasets. These tags are defined as follows:
\begin{itemize}
        \item \textbf{Cause}: Indicates the presence of causality
         \item \textbf{QFact}: Qualifies the causal chunk as quantitative (i.e., containing numerical entities like amounts)
         \item \textbf{Fact}: Qualifies the causal chunk as non-quantitative
         \item \textbf{Discard/Remove}: Indicates text that is not retained for the final datasets (non financial texts)
\end{itemize}

In addition, metrics related to \textbf{fact alignment} (i.e., trimming sentences according to preset priority rules in the annotation scheme) are also included to consistently reflect the preprocessing carried out at this step. All statistics are provided for the 3 datasets provided to participants: Trial, Practice, Evaluation as well as global statistics to present a general outlook on the overall annotation phase. The resulting statistics are collected in Table 1.

\begin{table}[h]
\begin{center}
\begin{tabular}{|l|r|r|r|r|r|r|}
\hline \bf Metric &  \bf Trial &  \bf Practice & \bf Evaluation & \bf Global \\ \hline
Total annotated files & 695 & 832 & 1878 & 3405  \\
\hline
Total sentences in files before definition of text sections & 25326 & 29381 & 74951 & 129658  \\
\hline
Total \textbf{Cause} tags in files & 657 & 1128 & 2244 & 4029 \\
\hline
Total \textbf{QFact} tags in files & 937 & 1824 & 2589 & 5350 \\
\hline
Total \textbf{Fact} tags in files & 449 &  999  & 2514 & 3962 \\
\hline
Total \textbf{Discard/Remove} tags in files & 1030 &  612  & 2462 & 4104 \\
\hline
Total files in review for \textbf{fact alignment} & 375 & 560  & 705 & 1640 \\
\hline
Total files modified in  \textbf{fact alignment} & 116 &  182  & 259 & 557 \\
\hline
Average causalities per file & 2.73 &  3.06 & 2.98 & 2.92 \\
\hline
Average offset of 2nd sentence in text sections & 137 &  139  & 141 & 139 \\
\hline
Average offset of 3rd sentence in text sections & 270 &  277  & 282 & 276 \\

\hline
Percentage of multi-sentenced text sections & 59.23 & 51.02 & 37.52 & 49.26 \\
\hline
\end{tabular}
\end{center}
\caption{\label{font-table} Global Distribution of Annotated files}
\end{table}

After fact alignment and inter annotator agreement (see Appendix A), a Trial and Training sets with Gold annotations were released, along with a blind Evaluation set for systems evaluation.

\section{Tasks}
Both substaks are intented as a pipeline. The first one aims at detecting if a text section contains a causal scheme (as defined in Appendix A.1), the second one aims at identifying cause and effect in a causal text section. Participants were allowed to concatenate and split the Trial and Practice datasets as they saw fit to train their system.

\subsection{Task1}
Task 1 is a binary classification task. The dataset consists of a sample of text sections labeled with 1 if the text section is considered containing a causal relation, 0 otherwise. The dataset is by nature unbalanced, as to reflect the proportion of causal sentences extracted from the original news, following the distribution displayed in Table 2.

\begin{table}[h]
\begin{center}
\begin{tabular}{|l|r|r|r|r|r|r|}
\hline \bf Metric &  \bf Trial &  \bf Practice & \bf Evaluation & \bf Global \\ \hline
Total number of text sections & 8580 & 13478 & 7386 & 29444  \\
\hline
Total number of causal text sections & 569 & 1010 & 567 & 2136  \\
\hline
Percentage of causal text sections & 6.63 & 7.49 & 7.68 & 7.24 \\
\hline
\end{tabular}
\end{center}
\caption{\label{font-table} Task 1 Distribution}
\end{table}

The trial and practice samples were provided to participants as csv files with headers \textit{Index; Text; Gold}. 
    \begin{itemize}
        \item Index: ID of the text section. Is a concatenation of  [file increment . text section index]
        \item Text: Text section extracted from a 2019 news article
        \item Gold: Gold Label provided from manual annotation
    \end{itemize}

\begin{table}[h]
\begin{center}
\resizebox{\textwidth}{!}{%
\begin{tabular}{ | c | l | c |}
      \hline
      \thead{Index} & \thead{Text} & \thead{Gold} \\
      \hline
      23.00005 & \makecell[l]{Electric vehicle manufacturers, components for the vehicles, batteries and producers for \\ charging infrastructure who invest over Rs 50 crore and create at least 50 jobs stand eligible \\ for total SGST (State GST) refund on their sales till end of calendar year 2030.} & 0 \\
      \hline
       23.00006	 & \makecell[l]{In case where SGST refund is not applicable, the state is offering a 15\% capital subsidy on \\ investments made in Tamil Nadu till end of 2025.}  & 1 \\
      \hline
\end{tabular}%
}
\end{center}
\caption{\label{font-table} Two examples from FinCausal Task 1 Corpus - Practice dataset}
\end{table}

\subsection{Task2}
The purpose of this task is to extract, in provided text sections, the chunks identifying the causal sequences and the chunks  describing the effects. 
The text sections correspond to the ones labeled as 1 in the Task 1 datasets, except in the blind Evaluation set.

The trial and practice samples were provided to participants as csv files with headers: \textit{Index; Text; Cause; Effect}
\begin{itemize}
    \item Index: ID of the text section.  Is a concatenation of  [file increment . text section index]
    \item Text: Text section extracted from a 2019 news article
    \item Cause: Chunk referencing the cause of an event (event or related object included)
    \item Effect: Chunk referencing the effect of the cause
\end{itemize}

Average statistics on the causes and effects chunks detected in the causal text sections are provided in Table 4. As explained in section 3, complex causal chains are considered during the annotation process, leading to one text section possibly containing multiple causes or effects. 

\begin{table}[h]
\begin{center}
\begin{tabular}{|l|r|r|r|r|r|r|}
\hline \bf Metric &  \bf Trial &  \bf Practice & \bf Evaluation & \bf Global Average \\ \hline
\makecell[l]{Average character length of causal chunks} & 113.73 & 109.13 & 112.48 & 111.78  \\
\hline
\makecell[l]{Average character length of effect chunks} & 107.79 & 104.78 & 99.66 & 104.08  \\
\hline
\makecell[l]{Total number of text sections } & 641 & 1109 & 638 & 796  \\
\hline

\makecell[l]{Total number of unicausal text sections} & 500 & 913 & 452 & 621.67  \\
\hline
\makecell[l]{Total number of multicausal text sections} & 141 & 196 & 186 & 174.33  \\
\hline
\end{tabular}
\end{center}
\caption{\label{font-table} Task 2 Distribution}
\end{table}

\begin{table}[h]

\begin{center}
\resizebox{\textwidth}{!}{%
\begin{tabular}{ | c | c | c | c |}
      \hline
      \thead{Index} & \thead{Text} & \thead{Cause} & \thead{Effect} \\
      \hline
      0009.00052.1	& \makecell[l]{Things got worse when the Wall came down. \\GDP fell 20\% between 1988 and 1993. \\ There were suddenly hundreds of thousands \\ of unemployed in a country that, \\ under Communism, had had full employment.}
      & \makecell[l]{ Things got worse when \\ the Wall came down.} & \makecell[l]{GDP fell 20\% between 1988 and 1993.} \\
      \hline
      0009.00052.2 &\makecell[l]{Things got worse when the Wall came down. \\ GDP fell 20\% between 1988 and 1993. \\There were suddenly hundreds of thousands \\ of unemployed in a country that,  \\ under Communism, had had full employment.} & \makecell[l]{ Things got worse when \\ the Wall came down.} & \makecell[l]{There were suddenly hundreds of thousands \\ of unemployed in a country that, \\ under Communism, had had full employment.} \\
      \hline
      23.00006	 & \makecell[l]{In case where SGST refund is not applicable,\\ the state is offering a 15\% capital subsidy \\ on investments made in Tamil Nadu \\ till end of 2025.} & \makecell[l]{In case where SGST refund \\ is not applicable} & \makecell[l]{the state is offering a 15\% capital \\ subsidy on investments made in \\ Tamil Nadu till end of 2025} \\

      \hline
\end{tabular}%
}
\end{center}
\caption{\label{font-table} Three examples from FinCausal Task 2 Corpus - Practice dataset}

\end{table}

\section{Evaluation}

A baseline was provided on the trial samples for both Tasks 1 and 2\footnote{\url{https://github.com/yseop/YseopLab/tree/develop/FNP_2020_FinCausal/baseline}}. 
Participating  systems  were ranked  on blind Evaluation datasets based on a weighted F1 score, recall, precision for Task 1, plus an additional Exact Match for Task 2.  Regarding official ranking, weighted metrics from the scikit-learn package\footnote{\url{https://scikit-learn.org/stable/modules/model_evaluation.html##multiclass-and-multilabel-classification}} were used for both Tasks, and the official evaluation script is available on Github\footnote{\url{https://github.com/yseop/YseopLab/tree/develop/FNP_2020_FinCausal/scoring}}.
Participating teams were allowed to submit as many runs as they wished, while only their highest score was withheld to represent them during evaluation. In addition, they were proposed to enhance their system in a post-evaluation phase\footnote{\url{https://competitions.codalab.org/competitions/25340}}. Only the scores validated during the evaluation phase of the competition are displayed below. Amongst the 13 participating teams, six choose to address Tasks 1 and 2 and one (ProsperAMnet) proposed an integrated pipeline for both.
Details on the methods and features used by different systems are provided in Table 8 for both Tasks. Noticeably, 7 teams plan to release the code associated to their system publicly.
 
\subsection{Task1}
Results for participating teams are provided in Table 6. Last line displays the baseline that had been provided for the task. The baseline was computed using the BERT-base-uncased language model\footnote{\url{https://huggingface.co/bert-base-multilingual-uncased}} and fine tuned on the Task data using the Hugging Face transformers library \cite{Wol:19}\footnote{\url{https://huggingface.co/transformers/model_doc/bert.html}}, on a GeForce GTX 1070 8Gb RAM GPU.
For Task 1, 6 participants out of 10 took advantage of large Transformers architectures \cite{Vas:17}  and fine-tuned their systems using the same library as the baseline. Four used Ensemble strategies to aggregate their results and enhance the robustness of their model. Additional strategies such as Data Augmentation and Oversampling are also proposed to work around the unbalanced nature of the data.
The best result in terms of weighted-averaged F1-score is achieved by the winning  team  LIORI  (97.75\%),  closely followed  by   UPB   and   ProsperAMNet with  F1  scores  of  97.55\%  and  97.23\%,  respectively.   The  top  five  systems all leveraged  Transformers architectures with associated language models features, evaluating at least on a fine-tuned BERT-base model and providing a comparison with similar models (BERT-large, RoBERTa, and specialized BERT such as FinBERT). The top 2 systems used Ensemble methods (See Table 8).  BERT-like systems weighted-F1 ratings are in range [97.75 , 95.78], whereas systems using more traditional Machine Learning models have scores in range [95.00 , 93.09], including systems using BERT-like embeddings in their processing.

\begin{table}[h]
\begin{center}
\begin{tabular}{|l|r|r|r|r|}
\hline \bf Team & \bf F1 Score & \bf Recall & \bf Precision  \\ \hline
LIORI  &	97.75 (1) &	97.77 (1) &	97.73 (1) \\
UPB  &	97.55 (2) &	97.59 (2) &	97.53 (2) \\
ProsperAMnet  &	97.23 (3) &	97.20 (3) &	97.28 (3) \\
FiNLP  &	96.99 (4) &	97.03 (4) &	96.96 (4) \\
DOMINO  &	96.12 (5) &	96.06 (5) &	96.19 (5) \\
IIT\_kgp 	 & 95.78 (6) &	95.83 (6) &	95.74 (6) \\
LangResearchLab\_NC  &	95.00 (7) &	94.92 (7) &	95.08 (7) \\
NITK NLP   &	94.35 (8)	& 94.87 (8) &	94.32 (8) \\
fraunhofer\_iais  &	94.29 (9) &	94.76 (9) &	94.20 (9) \\
ISIKUN   &	93.09 (10) &	94.33 (10) &	93.89 (10) \\
\hline
\hline
baseline & 95.23 & 95.21 & 95.26 \\

\hline
\end{tabular}
\end{center}
\caption{\label{font-table} Task 1 Results}
\end{table}

\subsection{Task2}

Results for Task 2 are provided in Table 7. Last line displays the baseline that has been provided for Task 2, computed with a CRF model using the pycrfsuite package\footnote{\url{https://python-crfsuite.readthedocs.io/en/latest/}}. One of the challenge of this task was to rebuilt the correct span of causal chunks, according to the annotation scheme. The baseline has been kept deliberately low as is does not take this specific problem into account, nor does it focuses on parameter-tuning strategies, though tuning examples are proposed with the code baseline.
All participants decided for sequence labelling strategies and used specific penalization methods and/or heuristics to work around the chunks reconstitution problem. The best performer in this subtask (NTUNLP) uses a BERT-CRF system and a Viterbi decoder for span optimization, achieving (94.72\%) weighted F1, closely followed by a BERT-SQUAD augmented system with heuristics for span  achieving 94.66\% F1 (Gbe).

\begin{table}[h]
\begin{center}
\begin{tabular}{|l|r|r|r|r|r|}
\hline \bf Team & \bf F1 Score & \bf Recall & \bf Precision  & \bf Exact match  \\ \hline
NTUNLPL  &	94.72 (1) &	94.70 (1) &	94.79 (1) &	82.45 (1) \\
GBe  &	94.66 (2) &	94.66 (2) &	94.67 (2) &	73.67 (2) \\
ProsperAMnet &	83.71 (3) &	83.63 (3) &	83.92 (3) &	70.38 (4) \\
LIORI  &	82.60 (4) &	82.80 (4) &	82.48 (4) &	70.53 (3) \\
DOMINO  &	79.60 (5) &	78.90 (5) &	81.90 (5) &	00.00 (7) \\
fraunhofer\_iais   &	76.00 (6) &	74.89 (7) &	79.95 (6) &	19.12 (5) \\
JDD  &	75.61 (7) &	75.57 (6) &	75.95 (7) &	00.00 (7) \\
UPB  &	73.10 (8) & 72.14 (8) &	75.61 (8) &	18.34 (6) \\
\hline
\hline
baseline & 51.06 & 51.74 & 50.99 & 11.11 \\

\hline
\end{tabular}
\end{center}
\caption{\label{font-table} Task 2 Results}
\end{table}

\begin{table}[h]
\begin{center}
\begin{tabular}{|c|c|c|c|c|c|c|c|c|c|c|}
\hline
\multirow{2}{*}{Team} & \multirow{2}{*}{F1} &  \multicolumn{9}{c|}{Techniques} \\ 
\cline{3-11}
& & ML & Neural & TF & Ens & AGM & RS & LM & WCS & HS \\
\hline
\multicolumn{11}{|c|}{Task 1} \\
\hline
LIORI &	97.75 & &  & X & X & &  & X & & \\
UPB &	97.55 & & & X & X & & & X & & \\
ProsperAMnet &	97.23 &  &  & X  & & &  & X &  & \\
FiNLP & 96.99 & &  & X & X & X & X & X & & \\
DOMINO & 96.12 & &  & X &  & & &  X & & \\
IIT\_kgp &	95.78 & &  & X &  & & & X & &\\
LangResearchLab\_NC	 & 95.00 & & X &  & & & X & X & X & \\
NITK NLP  &	94.35 & X & & & & & & & X & \\
fraunhofer\_iais &	94.29 & X &  & & X & & & & X &  \\
ISIKUN &	93.09 & X & & & & & & & X & \\
\hline
\multicolumn{11}{|c|}{Task 2} \\ \cline{1-11}
\hline
NTUNLPL &	94.72 & X &  & X & &  & & X & X & \\
GBe	& 94.66 &  &  & X & &  & & X & & X \\
ProsperAMnet &	83.71  &  &  & X & &  & & X &  & \\
LIORI &	82.60 & & & X & & & & X & & \\
DOMINO & 79.60  & &  & X & & & & X  & & X \\
fraunhofer\_iais &	76.00 & X & X & & & & & X & & X \\
JDD &	75.61 & X & X & & &  & & X & X & \\
UPB  &	73.10 & X & &  &  & & & X &  & X \\
\hline
\end{tabular}
\end{center}
\caption{\label{font-table} Approaches adopted by  the  participating  teams  in  Tasks  1  and  2.   ML refers  to  any  non-neural  machine  learning  technique such as XGBoost, SVM, etc.  Neural refers to any neural network architecture such as BILSTM, CNN, GRUs,etc, except Transformers.  TF refers to Transformers architecture.  Ens corresponds to Ensemble Learning method. RS is resampling method. HS implies some heuristics has been used in the final computation, mostly to adapt the span in Task 2. LM refers to any language model embedding features. WCS refers to Word, Character or Syntax based features.}
\end{table}

\section{Conclusion}
In this paper, we present the framework and the results for the FinCausal Shared Task.  In addition , we present the new FinCausal dataset built specifically for this shared task.  We plan to run similar shared tasks in the near future, possibly  with  some augmented data, in association with the FNP workshop.

\section*{Acknowledgements}
We would like to thank our dedicated annotators who contributed to the building the FinCausal Corpus:  Yagmur Ozturk, Minh Anh Nguyen, Aurélie Nomblot and Lilia Ait Ouarab, as well as the FNP Committee for their gracious support.

\begin{appendices}

\section{ Annotation scheme}
In this appendix, we provide detailed information on the concepts guiding the annotation.
The annotation process was iterative: Annotations were proposed on a BRAT annotation server\footnote{\url{https://brat.nlplab.org/installation.html}} by a first annotator then revised by two others until agreement. These agreement sessions were the opportunity to define and iterate on the following annotation scheme.

\subsection{Defining causatives}

A causal relationship involves the statement of a \textbf{cause} and its \textbf{effect}, meaning that two events or actors are related to each other with one triggering the other. 
We focused our annotation on text sections\footnote{We are using the term \textit{text section} since it could be a phrase, a sentence as well as a paragraph in which the cause and the effect are split in different sentences. For instance "Selling and  marketing  expenses decreased to \$1,500,000 in 2010. This was primarily attributable to employee-related actions and lower travel costs". However, in order to have a reproducible annotation process, we reduced the context to a paragraph of maximum three sentences.} that state causal relationships involving a quantified fact, which was necessary to reduce the complexity of the task. Table 9 displays the terms used in the context of the Shared Task.

\begin{table}[h]
\begin{center}
\resizebox{\textwidth}{!}{%
\begin{tabular}{|l|l|}

\hline
\multicolumn{2}{|c|}{FACT} \\
\hline
\hline
Empirical Fact & Past event, acknowledged  \\
Process & Concrete event in duration  \\
State of Affairs & In being situation (will become true or false) \\
Looking Forward Statement & Expectation (often a declaration made by CPY board member) \\
Hypothesis & Projection based on facts \\
\hline
\hline
\multicolumn{2}{|c|}{QUANTIFIED FACT (QFact)} \\ 
\hline
\hline
Explicit &  Has a direct connection to an explicit measure \\
Measurable & Measure is either a quantity or a number that can be precisely identified in a text section \\
Verifiable & Is a State of Affairs at least \\
\hline

\end{tabular}%
}
\end{center}
\caption{\label{font-table} Representation of events terminology}
\end{table}

In this scheme, an effect can only be a quantified fact. The cause can either be a fact or a quantified fact. The causality between these two elements can be implicit as well as explicitly stated with a triggering linguistic mark also called a connective. The place of these chunks in the text section can vary according to the connective used or simply according to the author's style.
\\

In order to delimit the process, the distance between a cause fact and an effect fact was restricted to \textbf{a 3-sentences distance}. In other words, we only annotated a causal relationship when there was a maximal gap of 1 untagged sentence between the two facts. For instance, in the text section \textit{\textless cause\textgreater Previous management sought to transform the company from a simple milk processor into a producer of value-added dairy products as it chased profits offshore\textless cause\textgreater.\textless effect\textgreater Among Fonterra's biggest missteps was the 2015 purchase of an 18.8 per cent stake in Chinese infant formula manufacturer Beingmate Baby \& Child Food for \$NZ755 million, just as the China market became hyper-competitive and demand slowed \textless effect\textgreater. Fonterra last month announced it would cut its Beingmate stake by selling shares after failing to find a buyer. Meanwhile, back home, Fonterra's share of the milk processing market dropped from 96 per cent in 2001 to 82 per cent currently, with consultants TDB Advisory expecting it to be about 75 per cent by 2021.} In this example, “the 2015 purchase of an 18.8 per cent stake in Chinese infant formula manufacturer Beingmate Baby \& Child Food for \$NZ755 million” was annotated because the cause and the effect have a 2-sentences distance. On the other hand, “Fonterra's share of the milk processing market dropped from 96 per cent in 2001 to 82 per cent currently” was not annotated because this effect is at a 4-sentences distance from the cause.

\subsection{Connectives}

A connective can be a verb, a preposition, a conjunction, an element of punctuation, or anything else, which explicitly introduces a causal relationship. Among those, there is a specific type of connective that is not taken into account in this Shared Task called lexical causative~\cite{Levin:15} . A lexical causative is a causal relationship stated through connectives (generally predicates) which, from a semantic point of view, also bear the effect of the cause. We will not consider those as causal references, since the effects are \textit{implied} in the connectives' definition. For instance in "The company raised its provisions by 5\% in 2018.", \textit{raise} is a lexical causative that can be glossed as \textit{The company caused the provisions to rise  by 5\%}.\par
Causal relationships can be introduced by other types of connectives in the identified text section. It is often rendered with the use of polysemous connectives which main function is not to introduce a causal relationship. For example, in this sentence: "Zhao found himself 60 million yuan indebted after losing 9,000 BTC in a single day (February 10, 2014)", the main function of the connective \textit{after} is to express a temporal relation between the two clauses. But we also have a causal relationship between them, since one triggers the other. \par
In the tagging process, the connectives involved in the causal relationship \textbf{were not annotated as part of the facts}. For example: \textless effect\textgreater \textit{Titan has acquired all of Core Gold's secured debt for \$US2.5 million}\textless effect\textgreater in order to \textless cause\textgreater \textit{ensure the long-term success of its assets.}\textless cause\textgreater. The only exception would be when the connective is inserted in the fact. In that case, the connective was annotated. For instance: \textless cause\textgreater\textit{On August 30, 2013, ST Yushun, in order to strengthen its competitive strength}\textless cause\textgreater, \textless effect\textgreater \textit{acquired a 100\% stake in ATV Technologies for 154 billion yuan}\textless effect\textgreater.

\subsection{Complex causal relationships}

In a text section, complex causal relationships can be rendered with conjoined relationships. A conjoined causal relationship can be one cause related to several effects, or one effect caused by several causes. This is often the case when the facts are not repeated and a conjunction is used as a link for the different effects or causes. This phenomenon can be also found in an implicit causal relationship and/or at sentence level. Here is an instance of a conjoined effect related to two causes: \textless cause\textgreater \textit{India's government slashed corporate taxes on Friday} \textless cause\textgreater \textit{,} \textless effect\textgreater \textit{giving a surprise \$20.5 billion break}\textless effect\textgreater \textless cause\textgreater \textit{aimed at reviving private investment and lifting growth from a six-year low that has caused job losses and fueled discontent in the countryside}\textless cause\textgreater. In the tagging process, they were all annotated as separate facts apart if a priority rule was to be taken into account.

\subsection{Priority rules}

The priority rules allow the annotation process of causal relationships to be more accurate and harmonious. \\
\par

\underline{First rule.} If a sentence contained \textbf{only one fact} (cause or effect), we \textbf{tagged the entire sentence} (even if it contains some noise or a connective). For instance: \textless cause\textgreater \textit{Hurricane Irma was the most powerful storm ever recorded in the Atlantic and one of the most powerful to hit land, Bonasia said.}\textless cause\textgreater \textless effect\textgreater It cause \$50 billion in damages.\textless cause\textgreater 
\\
\par \underline{Second rule.} The \textbf{annotation of sentence-to-sentence causal relationships is prioritized}. When the annotator had the choice between linking two full sentences together or subdividing a sentence, he chose the sentence-to-sentence annotation. To illustrate this point, let's look at the text section: "\textit{Finally, Seizert Capital Partners LLC increased its holdings in shares of BlackRock Enhanced Global Dividend Trust by 17.2\% during the second quarter. Seizert Capital Partners LLC now owns 138,020 shares of the financial services provider's stock valued at \$1,481,000 after acquiring an additional 20,223 shares in the last quarter.} In this text section, there are two causal relationships. The first one links “\textit{Seizert Capital Partners LLC increased its holdings in shares of BlackRock Enhanced Global Dividend Trust by 17.2\% during the second quarter}” and “\textit{Seizert Capital Partners LLC now owns 138,020 shares of the financial services provider's stock valued at \$1,481,000}”. Since the two facts are located into different sentences, we would have to annotate the full sentences each time (rule 1). The second causal relationship links “\textit{Seizert Capital Partners LLC now owns 138,020 shares of the financial services provider's stock valued at \$1,481,000}” and "\textit{acquiring an additional 20,223 shares in the last quarter}". Here, a sentence is subdivided. \par
Considering the priority of sentence-to-sentence annotation, the final annotation of this text section was: "\textless cause\textgreater \textit{Finally, Seizert Capital Partners LLC increased its holdings in shares of BlackRock Enhanced Global Dividend Trust by 17.2\% during the second quarter}\textless cause\textgreater . \textless effect\textgreater \textit{Seizert Capital Partners LLC now owns 138,020 shares of the financial services provider's stock valued at \$1,481,000 after acquiring an additional 20,223 shares in the last quarter.}\textless effect\textgreater ".\par

This rule also highlights the fact that \textbf{two different annotations cannot overlay}. It is impossible to annotate "\textit{acquiring an additional 20,223 shares in the last quarter}" and "\textit{Seizert Capital Partners LLC now owns 138,020 shares of the financial services provider's stock valued at \$1,481,000 after acquiring an additional 20,223 shares in the last quarter.}" because the same text segment would be part of two different annotations. \\

\underline{Third rule.} If a sentence contained \textbf{both a cause and an effect}, the \textbf{sentence was subdivided}. The spanning was realized so that the exact segments corresponding to the cause and the consequence were selected. For instance: \textit{This week's bad news comes from Rothbury, Michigan, where }
\textless cause\textgreater \textit{Barber Steel Foundry will close at the end of the year} \textless cause\textgreater, \textless effect\textgreater \textit{leaving 61 people unemployed}\textless effect\textgreater . However, in the dataset, the spans were extended in order to cover the entirety of the sentence. Only the connector, when located in between the cause and the effect, was left out of the extraction. As a result, in the final dataset we have: \textless cause\textgreater \textit{This week's bad news comes from Rothbury, Michigan, where Barber Steel Foundry will close at the end of the year} \textless cause\textgreater, \textless effect\textgreater \textit{leaving 61 people unemployed}\textless effect\textgreater . The spanning extension facilitate the consistency of the annotation process.\\
\par
\underline{Fourth rule.} If \textbf{two facts of the same type were located in the same sentence and were related to the same effect or cause}, then \textbf{we annotated these two facts as one unit}. For instance, in the text section "\textit{Thomas Cook's demise leaves its German operations hanging. More than 140,000 German holidaymakers have been impacted and tens of thousands of future travel bookings may not be honored.}", the cause fact is “\textit{Thomas Cook's demise}”. Since it was the only fact in the sentence, we annotated the full sentence as the cause (see priority rule number 1). The cause fact has two consequences: “\textit{More than 140,000 German holidaymakers have been impacted}” and “\textit{tens of thousands of future travel bookings may not be honored}”. Since both effect facts are in the same sentence and related to the same cause, we annotated the text section as follow: \textless cause\textgreater Thomas Cook's demise leaves its German operations hanging.\textless cause\textgreater \textless effect\textgreater More than 140,000 German holidaymakers have been impacted and tens of thousands of future travel bookings may not be honored\textless effect\textgreater .\par
This rule was also applied to the annotation of cause.s and effect.s inside a sentence. For instance: "\textless effect\textgreater \textit{Our total revenue decreased to \$31 million}\textless effect\textgreater due to \textless cause\textgreater \textit{decrease in orders from approximately \$91,000 to \$82,000, and a decrease in total buyers, which includes both new and repeat buyers from approximately 62,000 to 56,000.}\textless cause\textgreater ". The two causes were put together since they are related to the same effect.\par
This rule was only used in the two cases presented above. When more than two sentences were involved it was not taken into account. For example: "\textless cause\textgreater \textit{Let's say Shirley reduced her assets of \$165,000 through a gift of \$10,000 and pre-paying her funeral expenses for \$15,000.}\textless cause\textgreater \textless effect1\textgreater \textit{Her DAC would reduce from \$55 a day to \$43 a day (a saving of just over \$4,300 a year).}\textless effect1\textgreater \textless effect2\textgreater \textit{Her equivalent lump sum would reduce by almost \$88,000!}\textless effect2\textgreater ". Consequently, the same text section may appear twice in the release dataset.\\

\par
\underline{Fifth rule.} The annotation of \textbf{causal chains} inside a sentence. A segment of text that is a cause can also be the effect of another cause. For instance, the sentence "\textit{BHP emitted 14.7m tonnes of carbon dioxide equivalent emissions in its 2019 fiscal year, down from 16.5m tonnes the previous year due to greater use of renewable energy in Chile.}" contains three facts. "\textit{greater use of renewable energy in Chile} is the cause of \textit{down from 16.5m tonnes the previous year} which is also the cause of \textit{BHP emitted 14.7m tonnes of carbon dioxide equivalent emissions in its 2019 fiscal year}.\par
In that case, we \textbf{isolated the rightmost fact and tagged it according to its nature}. All \textbf{the remaining facts were gathered as one unit and annotated with the remaining tag}. In our example it gave the final annotation: "\textless effect\textgreater \textit{BHP emitted 14.7m tonnes of carbon dioxide equivalent emissions in its 2019 fiscal year, down from 16.5m tonnes the previous year}\textless effect\textgreater \textless cause\textgreater \textit{greater use of renewable energy in Chile}\textless cause\textgreater "

\subsection{Other annotation levels}
The cause or the effect can sometimes be found as pronouns, relative pronouns included. In that case, the reference (the antecedent) of the pronoun, is the extracted element. For instance, in the text: "The tax revenues decreased by 0.3\%, which was caused by fiscal decentralization reform." \textit{The tax revenue decreased by 0.3\%} corresponds to the effect and \textit{fiscal decentralization reform} is the cause. In some cases, the pronoun can be added to the opposite fact where the antecedent is.
\par
The role of a clause in a causal sentence can be ambiguous to identify. For example, it can be precarious to tell whether the clause corresponds to the cause, the means or the goal. If so, the sequence was annotated as the cause.\par
The ambiguity can also exist between two facts - which is the cause? which is the effect? In that case, when there was only one Qfact, the latter was annotated as the effect. When both facts were Qfacts, the annotation order was left to the annotator's appreciation. The annotator was encouraged to use reformulation in order to decide which fact was the cause and which fact was the effect.
\par
If the cause is in the middle of the effect or vice versa, the sentence is not annotated because of the conflict process. Here is an example: "The take-home pay after necessary deductions is S\$4,137." where \textit{after necessary deductions} is a cause inserted in the effect.
\par
We decided not to annotated causal relationships with structures identical to a calculation structure. For instance, in the text section "\textit{Google has 100K+ people and \$136B in revenue (2018), earning over \$1.3M per person.}", we considered that, since the quantified data in the effect fact is the result of a calculation based on the data present in the cause fact, there was no new information. Consequently, there was no need to annotate.
\par
Finally, dates are also to be included in the fact annotated if it is related to it and is placed next to it in the sentence.
\end{appendices}

\end{document}